\documentclass[letterpaper]{article} 
\usepackage{aaai2027} 
\usepackage[hyphens]{url} 
\usepackage{graphicx} 
\usepackage{natbib} 
\usepackage{caption} 
\usepackage{algorithm}
\usepackage{algorithmic}
\usepackage{booktabs}

\copyrighttext{arXiv submission. This version includes author names and affiliations.}

\title{Learning Compositional Meta-Routing for Agentic Workflows: An Executable Benchmark}
\author{
Natan Vidra\textsuperscript{\rm 1},
Alina Kapanova\textsuperscript{\rm 1,\rm 2},
Arun Kanhai\textsuperscript{\rm 1,\rm 3},
Spurthi Setty\textsuperscript{\rm 4}
}
\affiliations{
\textsuperscript{\rm 1}Anote AI\\
\textsuperscript{\rm 2}Cornell University\\
\textsuperscript{\rm 3}CUNY\\
\textsuperscript{\rm 4}Stevens Institute of Technology\\
nvidra@anote.ai, ak2765@cornell.edu, arun.kanhai55@qmail.cuny.edu, ssetty2@stevens.edu
}

\begin{document}

\maketitle

\begin{abstract}
Agentic systems must decide not only what answer to produce, but which reasoning and execution operations should precede it. A controller may answer directly, decompose a request, retrieve evidence, execute code, delegate to a specialist, or verify an intermediate result. Existing routing work largely selects model endpoints, retrieval depth, or tools in isolation. We introduce an executable benchmark and a budget-aware meta-router that composes heterogeneous operations from raw task text. The benchmark contains 216 training, 72 development, 108 held-out test, and 108 locked lexical-shift challenge tasks across data analysis, frozen-corpus research, and document processing. Outcomes are machine checked after operations execute. Independent regularized logistic heads predict operation probabilities from word and character features, are temperature-scaled on development data, and are greedily composed under route-cost and action-count budgets. On the held-out test, the learned policy achieves 100\% success versus 93.5\% for strong static and fixed workflows, with 43\% lower cost than the static policy; a matched learned one-shot router reaches 56.5\%. On the untouched challenge split, learned success falls to 75.9\% and trails static routing at 93.5\%, while remaining 49\% cheaper and exceeding one-shot routing by 34.3 points. The gap identifies lexical generalization, rather than route execution, as the principal limitation. These results establish a reproducible testbed and a bounded proof of concept, not evidence of live-LLM performance.
\end{abstract}


\section{Introduction}

Modern agentic applications combine language models with retrieval, code execution, external APIs, specialist agents, verification, and recovery logic. Before any component can help, a controller must choose whether to invoke it. This meta-decision affects accuracy, cost, latency, and exposure to component failure. A direct answer is efficient but may lack evidence or computation; indiscriminate tool use adds overhead; and a fixed workflow can fail when one component is unavailable.

Prior research demonstrates the value of adaptive retrieval \citep{jeong2024adaptiverag}, cost-aware model routing \citep{chen2024frugalgpt,ong2025routellm}, optimized function-call plans \citep{kim2024llmcompiler}, and reflective planning \citep{yao2023react,zhang2026spiral}. Agent benchmarks evaluate planning and tool execution \citep{liu2024agentbench,yao2024taubench}, but fewer studies isolate the policy that selects among \emph{different classes} of cognitive and execution operations. This leaves two questions entangled: whether a controller identifies relevant capabilities and whether composing several capabilities is better than selecting one.

We study that layer with an executable benchmark. Each task is presented to a policy only as raw text. A route may contain decomposition, retrieval, code execution, delegation, and verification before answering. Unlike our preliminary offline simulator, success is not sampled from annotated needs: operations transform task state, produce a candidate answer, and encounter deterministic failure conditions; an exact evaluator then checks the result. Labels are used only to train operation predictors and to analyze route quality.

Our contributions are:

\begin{itemize}
    \item an executable meta-routing benchmark with machine-checked data-analysis, research, and document-processing tasks, held-out templates, a locked lexical-shift challenge split, and component-outage cases;
    \item a calibrated raw-text router that predicts multiple operation types and composes them under explicit action-count and cost budgets;
    \item comparisons with direct, random, keyword, static-workload, fixed-agent, one-shot learned, and oracle policies on success, cost, local latency, budget compliance, and route quality; and
    \item paired statistical analysis and ablations separating composition, character-level generalization, calibration, and budget enforcement.
\end{itemize}

The central result is deliberately scoped. The learned router solves all 108 standard test tasks while using 1.76 normalized cost units, compared with 93.5\% success and 3.08 units for a strong static policy. Under lexical shift, however, it misses aggregate computation, multi-hop retrieval, and conflict verification, reaching only 75.9\%. The result shows both the value of operation composition and the fragility of lightweight text routing.

\section{Related Work}

\paragraph{Model and retrieval routing.}
FrugalGPT constructs model cascades to reduce inference cost, while RouteLLM learns to select between stronger and weaker models from preference data \citep{chen2024frugalgpt,ong2025routellm}. CP-Router uses uncertainty to choose between standard and long-reasoning models \citep{su2026cprouter}; ICL-Router represents endpoint capabilities through in-context vectors \citep{wang2026iclrouter}; and ZeroRouter uses a model-agnostic latent space to onboard unseen endpoints while optimizing accuracy, cost, and latency \citep{yan2026zerorouter}. Adaptive-RAG predicts question complexity and selects no retrieval, single-step retrieval, or iterative retrieval \citep{jeong2024adaptiverag}. These methods establish conditional routing as an operational optimization problem. Our decision variable is different: we compose heterogeneous operations around a model rather than choose a model or retrieval depth.

\paragraph{Tool learning and workflow construction.}
Toolformer learns when and how to call APIs \citep{schick2023toolformer}. API-Bank and ToolLLM provide resources for API planning, retrieval, and execution \citep{li2023apibank,qin2024toolllm}; AnyTool combines hierarchical retrieval with reflection after failed calls \citep{du2024anytool}; and RESTful-Llama studies an industry natural-language-to-API pipeline \citep{xu2024restful}. LLMCompiler exposes function dependencies for parallel execution and lower latency \citep{kim2024llmcompiler}, while Smurfs distributes tool planning across specialized agents and reduces context overhead \citep{chen2025smurfs}. These systems optimize tool-centered plans. We treat tool use as one optional operation beside decomposition, code, delegation, verification, and direct answering.

\paragraph{Planning and verification.}
ReAct interleaves reasoning with environmental action \citep{yao2023react}; SPIRAL separates planning, simulation, and critique inside reflective tree search \citep{zhang2026spiral}. ACPBench generates tasks with provably correct solutions from formal planning domains and finds uneven planning abilities across models \citep{kokel2025acpbench}. Our method is simpler: it predicts one bounded route before execution and does not perform search or observation-conditioned replanning. This simplicity makes the effect of route composition easier to isolate, at the cost of weaker adaptivity.

\paragraph{Agent evaluation and diagnosis.}
AgentBench, SWE-bench, and $\tau$-bench evaluate interactive, executable, or user-mediated behavior \citep{liu2024agentbench,jimenez2024swebench,yao2024taubench}. PlanningArena evaluates planning and tool selection at macro and micro levels \citep{zheng2025planningarena}; ToolSandbox adds stateful execution, intermediate milestones, and conversational evaluation \citep{lu2025toolsandbox}. AgentDiagnose measures trajectory competencies including decomposition, observation reading, verification, and backtracking \citep{ou2025agentdiagnose}. Our benchmark is smaller and less realistic, but provides controlled supervision and paired policy comparisons over several operation classes.

\paragraph{Novelty boundary.}
The contribution is not a new foundation model or general-purpose planner. It joins ideas normally evaluated separately: task-conditioned routing, multi-operation workflow construction, cost constraints, executable grading, and trajectory-level diagnosis. The matched one-shot learner sees the same text representation and supervision as the compositional router, so their difference isolates route composition more directly than comparisons against direct prompting alone.

\section{Problem Formulation}

Let a task be raw text $x$ with cost budget $B$. A route $r=(a_1,\ldots,a_m,A)$ terminates in answer action $A$ and draws support operations from

\begin{equation}
\mathcal{O}=\{D,T,C,G,V\},
\end{equation}

where $D$ is decomposition, $T$ retrieval/tool use, $C$ code execution, $G$ specialist delegation, and $V$ verification. Each operation has preregistered normalized cost $c(a)$. A route is feasible when $m\leq M$ and $\sum_{a\in r}c(a)\leq B$.

Executing $r$ on task $x$ produces candidate answer $\hat{y}$ and trace $z$. Success is $S(x,r)=1$ when $\hat{y}=y$ and $0$ otherwise under a task-specific exact evaluator. The policy objective is to maximize success under route constraints, while raw success, cost, latency, and failure type are reported separately. Operation labels $L(x)\subseteq\mathcal{O}$ identify a minimum valid route for training and route-quality analysis; the executor never converts label overlap directly into success.

\section{Executable Benchmark}

\subsection{Tasks and Operations}

The suite contains 504 tasks: 216 train, 72 development, 108 test, and 108 challenge examples, each balanced across three workloads. Prompt templates are disjoint across splits. The challenge prompts were locked after observing standard-test development behavior and use more distant paraphrases; no model or threshold was changed after their first run. Values, record identifiers, evidence entries, and document fields are generated deterministically from a fixed seed. Table~\ref{tab:tasks} summarizes the executable families.

\begin{table*}[t]
\centering
\small
\caption{Executable task families. Direct literal tasks occur in every workload. Minimum routes exclude the final answer action.}
\label{tab:tasks}
\begin{tabular}{p{0.15\textwidth}p{0.25\textwidth}p{0.15\textwidth}p{0.35\textwidth}}
\toprule
Workload & Task families & Minimum route & Machine-checked behavior \\
\midrule
Data analysis & Aggregate; filtered sum & $C$; $D,C$ & Compute numeric aggregates; applying code before decomposition sums the wrong subset \\
Research & Fact; multi-hop; conflicting sources; outage & $T$; $D,T$; $T,V$; $G$ & Retrieve frozen records, follow evidence chains, select newest evidence, or delegate when retrieval is unavailable \\
Document processing & Field extraction; invoice reconciliation; locale date; cross-field identifier & $T$; $T,C,V$; $T,G$; $D,T,V$ & Parse fields, recompute incorrect totals, normalize dates, and combine validated fields \\
All & Direct literal & none & Return an explicitly supplied literal without support operations \\
\bottomrule
\end{tabular}
\end{table*}

Operations manipulate an explicit state rather than receiving success credit from route labels. For example, retrieval over conflicting records initially returns the older entry; verification selects the record with the latest year. Invoice extraction initially exposes a stated but incorrect total; code computes the item sum and verification selects the recomputed value. In multi-hop research, retrieval without decomposition returns only the intermediate laboratory. Retrieval-outage tasks mark $T$ unavailable and provide the answer only through $G$. These mechanisms yield interpretable incorrect-answer, no-answer, and unavailable-component failures.

Execution is order sensitive. Decomposition writes task-specific subgoals or field requirements into state; retrieval and code consume those artifacts; verification can revise a candidate only when the prerequisite evidence or computation exists. If an unavailable operation is invoked, execution stops with a typed component failure. Otherwise, answering emits the current candidate or applies a direct-answer parser for literal tasks. Thus an irrelevant extra operation may add cost without changing the answer, while a missing or misordered prerequisite changes the actual artifact presented to the evaluator.

The benchmark design separates four sources of difficulty. \emph{Selection} tasks require one non-answer operation. \emph{Composition} tasks require two or three dependent operations. \emph{Conflict} tasks provide a plausible but wrong intermediate candidate that verification must correct. \emph{Availability} tasks invalidate the usual retrieval path and require delegation. This taxonomy supports failure analysis beyond aggregate accuracy and prevents a direct policy from succeeding through answer-format heuristics on nonliteral tasks.

The operation costs are $.45$ for decomposition, $1.00$ for retrieval, $1.15$ for code, $1.40$ for delegation, $.65$ for verification, and $.30$ for answering. Every task uses budget $B=4.5$. Costs represent relative component usage, not dollars or tokens.

\subsection{Raw-Text Operation Model}

The router receives only the prompt text. We construct binary word unigrams, word bigrams, and character 3--5-gram features $\phi(x)$. For each operation $o$, a regularized logistic head estimates

\begin{equation}
p_o(x)=\sigma(w_o^\top\phi(x)+b_o).
\end{equation}

Heads minimize class-balanced binary cross entropy with $\ell_2$ regularization on the training split. Optimization uses 500 full-batch gradient steps with a decaying learning rate. We temperature-scale each head on development data by minimizing Brier score over a fixed temperature grid. Development Brier scores range from $.00004$ for delegation to $.102$ for retrieval.

For operation $o$, the training objective is

\begin{equation}
\mathcal{L}_o=\mathrm{WBCE}(y,p_o(x))+\lambda\|w_o\|_2^2,
\end{equation}

where weighted binary cross entropy (WBCE) gives positive and negative examples equal total weight. We use learning rate $.35/\sqrt{1+e/50}$ at epoch $e$, $\lambda=.002$, and minimum document frequency 2. The resulting representation has 3,349 features. Training all five heads takes approximately $.12$ seconds on the reported CPU. Temperature selection chooses $.5$ for every head; the no-calibration ablation below tests whether this affects routes.

The route threshold $\tau$ is selected from $\{.30,.35,\ldots,.70\}$ on development labels, prioritizing exact-route accuracy, then action F1, then lower cost. This yields $\tau=.40$. Candidate operations with $p_o(x)\geq\tau$ are ranked by $(p_o-\tau)/c(o)$ and greedily added while respecting $M=3$ and $B$. Operations are placed in canonical order $D,T,C,G,V$, followed by $A$. Algorithm~\ref{alg:router} gives the procedure.

\begin{algorithm}[t]
\caption{Budget-Aware Text Meta-Routing}
\label{alg:router}
\begin{algorithmic}[1]
\REQUIRE text $x$, probabilities $p_o$, threshold $\tau$, budget $B$
\STATE $Q\leftarrow\{o:p_o(x)\geq\tau\}$
\STATE sort $Q$ by $(p_o(x)-\tau)/c(o)$ descending
\STATE $R\leftarrow[\,]$; $k\leftarrow c(A)$
\FOR{$o$ in $Q$}
    \IF{$|R|<3$ and $k+c(o)\leq B$}
        \STATE append $o$ to $R$; $k\leftarrow k+c(o)$
    \ENDIF
\ENDFOR
\STATE reorder $R$ as $D,T,C,G,V$
\RETURN $R$ followed by $A$
\end{algorithmic}
\end{algorithm}

\subsection{Baselines}

We compare seven alternatives. \emph{Direct} always answers. \emph{Random} selects zero to three operations deterministically from the task identifier. \emph{Keyword} uses raw-text rules. \emph{Static workload} uses one route per workload. \emph{Fixed agent} always executes all five support operations. \emph{Learned one-shot} uses the same calibrated model but selects at most its highest-probability operation. \emph{Oracle} executes the annotated minimum route and serves as an upper bound. Static and fixed policies are deliberately strong: they solve all tasks except research outages.

\section{Experimental Protocol}

We fit on train, calibrate and select $\tau$ on development, and report both standard test and the subsequently locked challenge split. No standard-test or challenge labels are used for fitting or threshold selection. Every policy runs every task in each evaluation split. We report machine-checked success, 2,000-sample task-bootstrap 95\% intervals, normalized route cost, local wall-clock route-plus-execution latency, budget compliance, exact-route match, and micro action precision/recall/F1. Paired success differences use task bootstrap intervals and an exact two-sided sign test over discordant outcomes.

Experiments ran CPU-only on an Apple M5 system with 10 cores and 16 GB memory under macOS, Python 3.12.13, and NumPy 2.5.0. Routing and execution are repeated five times per task solely to stabilize sub-millisecond timing, then averaged. The local operations do not call an LLM or network service; latency therefore measures implementation overhead, not deployment latency.

\section{Results}

\subsection{Main Comparison}

\begin{table}[t]
\centering
\small
\caption{Held-out executable results ($n=108$ per policy). CI is the bootstrap interval for success. Latency is local milliseconds.}
\label{tab:main}
\begin{tabular}{lrrrr}
\toprule
Policy & Success & 95\% CI & Cost & F1 \\
\midrule
Oracle & 1.000 & [1.000,1.000] & 1.50 & 1.000 \\
Learned budget & \textbf{1.000} & [1.000,1.000] & 1.76 & .910 \\
Static workload & .935 & [.880,.972] & 3.08 & .535 \\
Fixed agent & .935 & [.880,.972] & 4.95 & .414 \\
Keyword & .583 & [.491,.676] & 1.13 & .810 \\
Learned one-shot & .565 & [.472,.657] & 1.36 & .643 \\
Random & .407 & [.315,.500] & 1.68 & .303 \\
Direct & .259 & [.185,.352] & .30 & .000 \\
\bottomrule
\end{tabular}
\end{table}

The learned budget policy solves all 108 tasks. Relative to static workload routing, it improves success by 6.48 points (paired 95\% CI [2.78,11.11], exact sign $p=.0156$) and reduces mean cost by 1.32 units, or 43.0\%. The difference consists of seven retrieval-outage tasks solved only by learned delegation. Relative to learned one-shot routing, composition improves success by 43.52 points ([34.26,52.78], $p<10^{-13}$) for $.40$ additional cost. This is the strongest evidence that multiple predicted operations, not text supervision alone, drive task completion.

The learned policy has 100\% budget compliance versus 66.7\% for static routing and 0\% for the fixed agent. Its mean local latency is $.17$ ms, compared with less than $.01$ ms for rule policies; this difference is measurable but operationally negligible relative to model or tool calls.

\subsection{Locked Challenge Split}

\begin{table}[t]
\centering
\small
\caption{Results on 108 locked lexical-shift challenge tasks.}
\label{tab:challenge}
\begin{tabular}{lrrr}
\toprule
Policy & Success & Cost & Action F1 \\
\midrule
Oracle & 1.000 & 1.50 & 1.000 \\
Static workload & .935 & 3.08 & .535 \\
Fixed agent & .935 & 4.95 & .414 \\
Learned budget & .759 & 1.56 & .810 \\
Learned one-shot & .417 & 1.19 & .549 \\
Random & .380 & 1.87 & .304 \\
Keyword & .324 & .54 & .331 \\
Direct & .259 & .30 & .000 \\
\bottomrule
\end{tabular}
\end{table}

\begin{figure*}[t]
\centering
\includegraphics[width=0.92\textwidth]{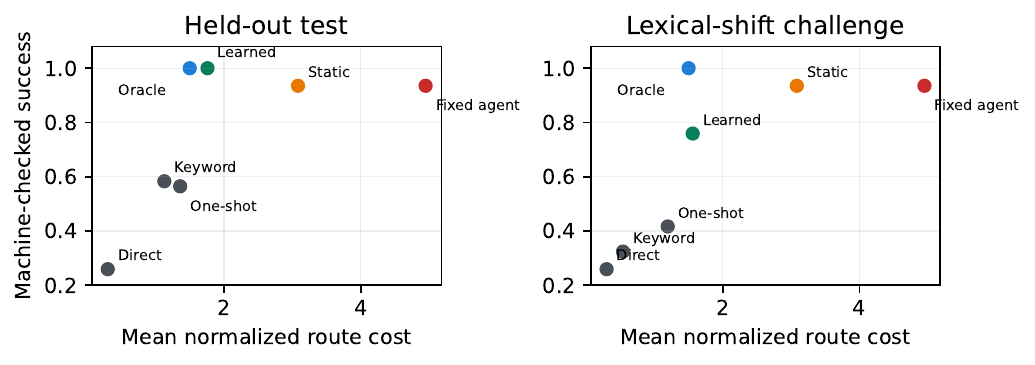}
\caption{Machine-checked success versus route cost. The learned router matches oracle success on the standard test but loses to static routing under lexical shift, while remaining substantially cheaper.}
\label{fig:tradeoffs}
\end{figure*}

The challenge reverses the comparison with static routing. Learned success is 17.59 points lower (paired 95\% CI [$-27.78$,$-7.41$], $p=.0013$), although cost is 1.52 units, or 49.4\%, lower. The learned policy still exceeds one-shot routing by 34.26 points ([25.93,44.44], $p<10^{-10}$). All 26 learned failures cluster in three families: 12 aggregate tasks, seven multi-hop research tasks, and seven conflicting-source tasks. Thus the executor and route composition remain effective when the required operations are recalled; the failure is operation prediction under paraphrase. Figure~\ref{fig:tradeoffs} contrasts the two splits.

\subsection{Route Quality and Ablations}

Although learned and oracle success are equal, their routes differ. The learned policy has .741 exact-route match, .834 action precision, 1.000 recall, and .910 F1. It invokes retrieval on 77 tasks versus 49 for the oracle, explaining its $.26$ higher cost. Extra operations do not change answers in the current executor, but they expose an efficiency gap hidden by success.

\begin{table}[t]
\centering
\small
\caption{Learned-router ablations on the same test tasks.}
\label{tab:ablations}
\begin{tabular}{lrrr}
\toprule
Condition & Success & Cost & Action F1 \\
\midrule
Full & \textbf{1.000} & 1.76 & .910 \\
No calibration & 1.000 & 1.76 & .910 \\
No budget enforcement & 1.000 & 1.76 & .910 \\
Word features only & .870 & 1.37 & \textbf{.948} \\
Single operation & .565 & 1.36 & .643 \\
\bottomrule
\end{tabular}
\end{table}

Removing character features lowers success by 12.96 points despite increasing action F1. The word-only model is more precise but misses operations under held-out phrasing; character features favor recall. Restricting the model to one operation produces the largest loss. Calibration and budget ablations are identical to the full policy because they do not alter threshold crossings or activate the budget boundary on this split. They should not be credited for the main gain under the tested regime.

\subsection{Sensitivity and Failure Analysis}

\begin{table}[t]
\centering
\small
\caption{Threshold sensitivity without refitting. Costs are mean normalized route costs.}
\label{tab:sensitivity}
\begin{tabular}{lrrrr}
\toprule
$\tau$ & Test & Cost & Challenge & Cost \\
\midrule
.30 & 1.000 & 1.76 & .759 & 1.56 \\
.40 & 1.000 & 1.76 & .759 & 1.56 \\
.50 & 1.000 & 1.76 & .722 & 1.52 \\
.60 & 1.000 & 1.50 & .694 & 1.50 \\
.65 & .963 & 1.48 & .639 & 1.17 \\
.70 & .926 & 1.45 & .546 & 1.09 \\
\bottomrule
\end{tabular}
\end{table}

Table~\ref{tab:sensitivity} shows a stable region from $\tau=.30$ to $.45$. Larger thresholds trade cost for recall and harm the challenge split first. At $\tau=.60$, standard success remains perfect and routes match the oracle exactly, but challenge success is only 69.4\%. Choosing a threshold solely from the standard distribution therefore understates generalization risk.

The challenge failures are sharply attributable. On aggregate prompts, mean code probability is below threshold (a representative task has $p_C=.192$). Multi-hop prompts retain retrieval ($p_T=.882$) but miss decomposition ($p_D=.051$), returning the intermediate laboratory. Conflicting-source prompts retain retrieval ($p_T=.985$) but miss verification ($p_V=.009$), returning stale evidence. In contrast, the outage paraphrase still selects delegation ($p_G=.675$), and invoice reconciliation retains retrieval, code, and verification probabilities above $.67$. These traces show that failure is not uniform semantic collapse; it is operation-specific recall under paraphrase.

We also refit with 36, 72, 108, 144, and 216 training tasks. Standard and challenge success are unchanged at every size. Because each family repeats one training template with different values, this flat curve indicates redundancy, not exceptional sample efficiency. Broader template and domain diversity is necessary before drawing scaling conclusions.

\section{Discussion}

\subsection{Interpretation and Novelty}

Three conclusions follow within the benchmark. First, static workflows remain strong when workload identity predicts the needed route; they waste cost on easy tasks and fail under component outages, but are more robust to paraphrase than the learned lexical router. Second, predicting a single operation is insufficient for tasks such as filtered computation, multi-hop retrieval, invoice reconciliation, and locale normalization on both evaluation splits. Third, standard-test success alone overstates router quality: the learned policy matches the oracle on answers while selecting 28 unnecessary retrieval operations and then loses 24.1 success points under lexical shift.

The method inherits task conditioning from adaptive retrieval, cost awareness from model routing, and modular operations from agent planning. Its novel unit of control is the heterogeneous operation route. Endpoint routers decide which model should answer; retrieval routers decide how much evidence to acquire; tool planners arrange calls within a tool-centered workflow. Our policy decides which kinds of reasoning and execution should constitute the workflow before those lower-level mechanisms run. The executable suite and matched one-shot control make this distinction empirically testable.

\subsection{Deployment Implications}

A practical system could use this router above existing model and tool routers. A low-cost gate predicts whether direct answering is sufficient; a multi-label policy then proposes support operations under cost and latency budgets; endpoint selection occurs inside each operation. Typed traces support per-component reliability estimates, budget audits, and fallback rules. The outage cases illustrate why route choice should depend on component availability rather than only task category.

The present classifier is intentionally lightweight and interpretable. A deployment router should replace lexical features with representations trained across broader tasks, condition on execution history and component health, and replan after observations. A constrained contextual bandit or Markov decision process could optimize success and cost online, while retaining static routes as fallbacks when uncertainty is high.

The challenge result suggests a concrete hybrid. When calibrated operation probabilities are diffuse or fall near threshold, the controller can back off to a workload route rather than answer with an incomplete composition. On the current challenge set, static routing covers the aggregate, multi-hop, and conflict families missed by the learner, while learned delegation covers the static policy's outage failures. A learned confidence gate between these policies is a natural next baseline, but we leave it unevaluated to avoid tuning on the locked challenge split.

\subsection{Limitations and Threats to Validity}

\paragraph{Construct validity.}
Machine-checked answers are stronger than sampled success, but the operations are deterministic Python components rather than LLMs, sandboxes, or network tools. Normalized route costs are design constants, not tokens or dollars. Local sub-millisecond latency should not be extrapolated to deployed agents.

\paragraph{Internal validity.}
Minimum-route labels and executor semantics are designed together. The executor does not directly score label overlap, but task families still encode the authors' assumptions about which operations should help. Perfect learned success reflects complete recall on a finite templated suite, not general intelligence. The threshold and temperatures are selected on only 72 development tasks.

\paragraph{External validity.}
The locked challenge split exposes surface-form fragility but does not test new domains, languages, tools, or long interactive horizons. PlanningArena and ToolSandbox contain richer stateful interactions \citep{zheng2025planningarena,lu2025toolsandbox}; SWE-bench and $\tau$-bench provide more realistic executable and user-mediated outcomes \citep{jimenez2024swebench,yao2024taubench}. Evaluation on those settings, semantic encoders, multiple model families, and adversarial distribution shifts is required.

\paragraph{Benchmark-development validity.}
The standard test influenced benchmark and feature development before the challenge split was introduced. We therefore treat standard-test performance as an in-benchmark result, not a pristine confirmatory estimate. The challenge templates were subsequently locked and the first-run outcome was retained without model changes, but they were still authored with knowledge of the benchmark ontology. An independently constructed task suite is needed for confirmatory evaluation.

\paragraph{Statistical validity.}
Task bootstrap intervals quantify variation over this finite suite. The 100\% success interval is degenerate and does not estimate performance on an external task population. Exact sign tests cover paired outcome differences but do not correct across every exploratory comparison. Future work should preregister a larger benchmark and report family-stratified uncertainty.

\section*{Ethical Statement}

The benchmark contains synthetic records, documents, and numerical tasks; it uses no personal, proprietary, defense, or human-subject data and performs no external actions. Deployed meta-routing can nevertheless create privacy, security, and accountability risks when data are sent to external services, code is executed, or decisions are delegated. Practical systems should use sandboxing, allowlisted tools, data minimization, hard budgets, audit logs, and human escalation for high-impact tasks. Benchmark success must not be interpreted as evidence of safety or suitability for autonomous consequential decisions.


\bibliography{references}

\end{document}